%% file: main.tex
\documentclass[a4paper, 10pt, conference]{ieeeconf}      % Use this line for a4 paper

\IEEEoverridecommandlockouts

\usepackage{graphicx} % for pdf, bitmapped graphics files
\usepackage{subcaption}
\usepackage{caption}

\usepackage{epsfig} % for postscript graphics files
\usepackage{mathptmx} % assumes new font selection scheme installed
\usepackage{comment}

\usepackage{amsmath,amssymb,amsfonts}
\usepackage{algorithmic}

\usepackage{pgfplots}

\usepackage{tabularx}
\usepackage{url}
\usepackage{csquotes}

\usepackage{cleveref}
\usepackage{etoolbox}
\makeatletter

\usepackage{accents}

\usepackage{lipsum}
\usepackage{xcolor}
\usepackage{multirow}

\usepackage[a4paper,bindingoffset=0.2in,%
left=0.556in,right=0.556in,top=0.944in,bottom=0.944in,%
footskip=.25in]{geometry}

\title{
\LARGE \bf neuROSym: Deployment and Evaluation of a ROS-based Neuro-Symbolic Model for Human Motion Prediction

\author{Sariah Mghames$^1$, Luca Castri$^1$, Marc Hanheide$^1$, Nicola Bellotto$^{1,2}$
\thanks{\hspace{-3mm}\textsuperscript{1}School of Computer Science, University of Lincoln, UK.\newline
\textsuperscript{2}Dept. of Information Engineering, University of Padua, Italy.\newline
This project has received funding from the EU's Horizon 2020 Research and Innovation programme under grant agreement No 101017274}
}
}
\begin{document}
\maketitle
\thispagestyle{empty}
\pagestyle{empty}

%\iffalse
\input{abstract}

\input{intro}

\input{literature}

\input{approach}

\input{experiments}  
\input{conclusion}

%\section*{Acknowledgement}

\bibliographystyle{IEEEtran}
\bibliography{IEEEabrv,references}

\end{document}

%% file: abstract.tex
\begin{abstract}
Autonomous mobile robots can rely on several human motion detection and prediction systems for safe and efficient navigation in human environments, but the underline model architectures can have different impacts on the trustworthiness of the robot in the real world. Among existing solutions for context-aware human motion prediction, some approaches have shown the benefit of integrating symbolic knowledge with state-of-the-art neural networks. In particular, a recent neuro-symbolic architecture~(NeuroSyM) has successfully embedded context with a Qualitative Trajectory Calculus~(QTC) for spatial interactions representation. This work achieved better performance than neural-only baseline architectures on offline datasets. In this paper, we extend the original architecture to provide \emph{neuROSym}, a ROS package for robot deployment in real-world scenarios, which can run, visualise, and evaluate previous neural-only and neuro-symbolic models for motion prediction online. We evaluated these models, NeuroSyM and a baseline SGAN, on a TIAGo robot in two scenarios with different human motion patterns. We assessed accuracy and runtime performance of the prediction models, showing a general improvement in case our neuro-symbolic architecture is used. We make the \emph{neuROSym} package\footnote{\url{https://github.com/sariahmghames/neuROSym}} publicly available to the robotics community.

\end{abstract} 

%\lipsum[1]

%% file: intro.tex
\section{Introduction}  \label{sec:intro}
The integration of autonomous mobile robots in logistics, transportation, and healthcare, is rapidly increasing, as is user trust in the technologies used to build them. One key requirement for mobile robots' trustworthiness is the ability to navigate safely among humans, in addition to other capabilities such as intelligent interaction and successful task completion. Safe navigation usually requires the detection and prediction of human motion. This is necessary not only for autonomous navigation, but also for action and intent recognition, anomaly detection, and other tasks. Existing methods for human motion detection and prediction can be divided into context-agnostic and (static or dynamic) context-aware. Taking context into account (e.g. knowing whether the robot is in a warehouse or a supermarket) can have a significant impact on the accuracy of the motion prediction system.

\begin{figure}[t]\centering
\includegraphics[scale=0.07]{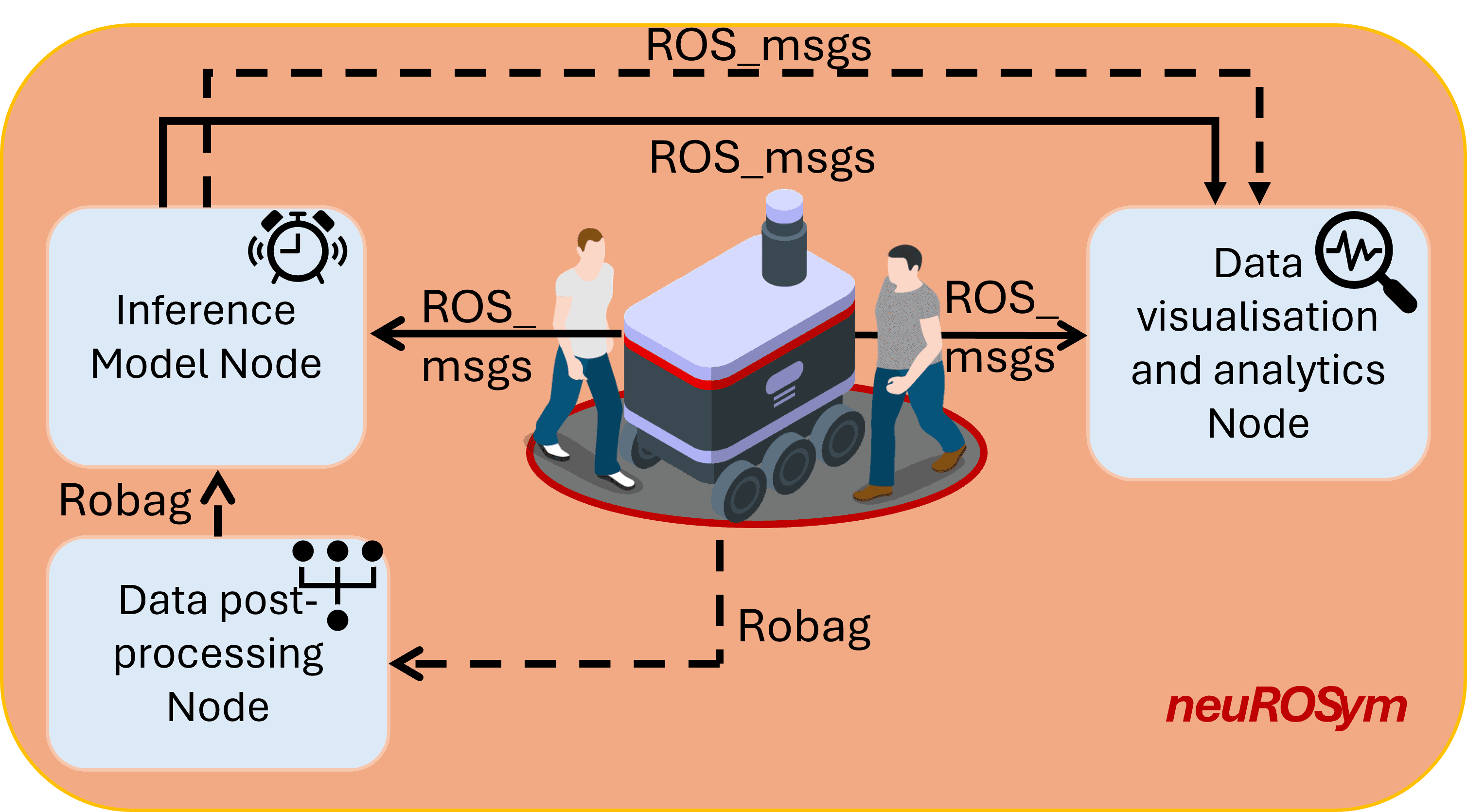}
\caption{Deployment of \emph{neuROSym} for online and context-aware human motion prediction, with real-time visualisation. The three blocks are ROS nodes, while the filled arrows and the dashed ones represent the online and offline inference, respectively. Each arrows label indicates the type of messages published and/or subscribed to by each node.} \label{fig:intro}
\end{figure}

%  Context reasoning 
Context reasoning can include human-human and/or human-objects spatial interactions. The latter can be described by quantitative or qualitative representations.
In the quantitative approach, interactions are typically embeddings of absolute or relative agents pose in a neural network model. The authors  in~\cite{chen2019crowd} have jointly modeled human-robot and human-human interactions in a deep reinforcement learning framework for mobile robot navigation.
In~\cite{pokle2019deep}, instead, the authors learn an optimal local trajectory from a global plan by fusing human trajectories, LiDAR features, global path, and odometry features in particular attention layer. 
Context-awareness methods have also been proposed to deal with the challenges faced by the long-term prediction of single human motion~\cite{Alahi2016,lisotto2019social,gupta2018social,mangalam2020not,liang2019peeking,Cao2020}.

Qualitative representations of spatial interactions, though intuitive and computationally efficient, are less explored for context-aware human motion prediction. Recently, a new approach has been proposed that exploits a qualitative representation of spatial interactions to improve human motion prediction~\cite{mghames2023neuro}. This consists in a neuro-symbolic model that has been proved to be effective in predicting human motion trajectories. The symbolic part is indeed a qualitative representation of spatial interactions between pairs of agents using the so-called Qualitative Trajectory Calculus~(QTC)~\cite{delafontaine2012qualitative,bellotto2013qualitative}. QTC-based models of moving agent pairs can be described by different combinations of QTC symbols, which depends on properties of the interaction such as relative distance changes (i.e moving towards/away), velocity (i.e moving faster/slower), and orientation (i.e. moving to the left/right side).

% Deployment 
However, deploying and validating these methods for context-aware human motion prediction on real-world robot domains has been only partly explored~\cite{gupta2018social,mghames2023neuro,liang2019peeking,lee2017desire,mangalam2020not}. While offline experimental evaluation is essential for model comparison and selection, the actual performance of any chosen architecture can only be validated during the deployment phase, since real-world problems such as domain-shift and inference time can significantly affect the human prediction system.
Based on the above considerations, and extending our previous NeuroSyM model for context-aware human motion prediction~\cite{mghames2023neuro}, the main contribution of this paper is two-fold: 
\begin{itemize}
    \item a new ROS package for online human motion prediction and visualisation, called~\emph{neuROSym}, which is publicly available 
    %for the robotics community 
    and includes the three blocks (inference, post-processing, and visualisation) depicted in Fig.~\ref{fig:intro};
    \item a performance evaluation of the package on a real-world robot scenario with two different experimental settings: (i)~people moving in the same directions, and
    %evolving from the tendency to avoid crossing other agents path, 
    (ii)~people crossing each other's paths. 
    %evolving from the tendency to cross other agents path.
\end{itemize}

The remainder of the paper is as follows: Sec.~\ref{sec:lit} presents an overview of the related works; Sec.~\ref{sec:appr} explains the approach adopted for human-context reasoning and deployment; Sec.~\ref{sec:exp} illustrates and discusses the experimental results; finally, Sec.~\ref{sec:conc} concludes by summarising the main outcomes and suggesting future research directions.

%% file: literature.tex
\section{Related Work} \label{sec:lit}

%\textbf{Human-Objects interactions modeling:}
\textbf{Context-aware human motion prediction.} 
Among the existing works in the area of context-aware human motion prediction~\cite{rudenko2020human}, some incorporate spatio-temporal dependencies between interactions~\cite{tao2020dynamic,yu2020spatio,huang2019stgat}, while others consider spatial relations only~\cite{Alahi2016,lisotto2019social,gupta2018social,mangalam2020not,liang2019peeking,Cao2020,lee2017desire,bisagno2021embedding}. These can be further grouped in solutions that take into account static context~\cite{Cao2020}, dynamic context~\cite{Alahi2016,gupta2018social,mangalam2020not,yu2020spatio,huang2019stgat}, or both~\cite{tao2020dynamic,lisotto2019social,liang2019peeking,lee2017desire,bisagno2021embedding}.

Two of the most popular architectures for human motion prediction are Social-LSTM~(S-LSTM)~\cite{Alahi2016} and Social Generative Adversarial Network~(SGAN)~\cite{gupta2018social}. Both use a spatially-aware pooling mechanism for incorporating the hidden states of nearby agents as a way to overcome the problem of variable and (potentially) large number of people in the scene. SGAN, however, is generally better than S-LSTM in terms of accuracy and time complexity, since it avoids grid-based pooling. SGAN features also lower time complexity and number of parameters compared to the Spatial-Temporal Graph Attention~(STGAT) network~\cite{huang2019stgat}. Recently, we propsed a new neuro-symbolic approach~\cite{mghames2023neuro} for context-aware human motion prediction, called NeuroSyM, which showed higher prediction accuracy on public datasets compared to SGAN.  Other promising approaches have also demonstrated to improve networks performance, like the endpoint conditioned trajectories prediction in~\cite{mangalam2020not}, the combined future activities and location prediction in~\cite{liang2019peeking}, and the dynamic and static context-aware motion predictor in~\cite{tao2020dynamic}.
%which integrates dynamic interactions between agents in a Social-aware Context  Module (SCM), whereas the static context is incorporated in a latent space with a semantic scene mapping.

In this paper, we focus on the dynamic aspect of context, since it is typically the most challenging part for a mobile robot. We study in particular the real-world performance in human motion prediction of NeuroSyM against an SGAN baseline, implemented in a common ROS-based software framework for robot deployment.
%Other advanced architectures, featuring static- rather than dynamic-context awareness~\cite{liang2019peeking,mangalam2020not} could be integrated in future work.}

\textbf{Human-human interaction modeling:}
The methods to represent the interactions of nearby agents can be divided into one-to-one modeling and crowd modeling~\cite{Ijaz2015,hedayati2020reform,thompson2021conversational}.
One-to-one interactions can be described by quantitative or qualitative representations. While the former have recently been modeled by multi-layer perceptrons embedding relative positions or velocities of agent pairs~\cite{chen2019crowd,gupta2018social,Alahi2016}, qualitative approaches use symbolic representations, for example QTC-based~\cite{bellotto2013qualitative,hanheide2012analysis,dondrup2014probabilistic}.
%In~\cite{dondrup2014probabilistic}, the use of qualitative rather than quantitative representations for analysing human-robot spatial interactions (HRSI) was motivated by the need of a more intuitive understanding of the observed interactions.
Similar models were used in~\cite{dondrup2016qualitative} to implement human-aware robot navigation strategies, where the prediction of interactions was limited to a Bayesian temporal model of single human-robot pairs, without taking into account nearby static or dynamic objects.

In our study, we consider one-to-one (i.e. pairwise) interactions
%due to the different nature of interactions that may occur in the neighbourhood of a single agent. We leverage on
through our previous NeuroSyM prediction system~\cite{mghames2023neuro}, which exploits QTC relations to weight the quantitative embedding of spatial interactions. However, we consider all the pairwise interactions in the neighbourhood, not just a single human-robot one.
%taking inspiration from hybrid approaches of crowd modeling for crowdy scenes.

\textbf{Human motion prediction deployment.}
Most of the research on human motion prediction is evaluated on public datasets.
In~\cite{zhao2020experimental}, the authors compared four different types of online human motion prediction for safe and efficient human-robot collaboration. Their models use linear regression and neural networks, with or without parameters adaptation. The authors showed that adaptable prediction models parameterized by neural networks achieved the best performance. They did not consider context-aware and long-term predictions though.

In~\cite{luber2010}, human motion prediction with Social Force Models~(SFMs) was exploited for real-world people tracking.
The work in~\cite{gui2018teaching} proposes a GAN-based solution that teaches the robot to mimic and predict human motion, but with a focus on actions rather than walking trajectories, and without taking into account context. In~\cite{6907734} instead, the authors validated a probabilistic framework, based on SFM and intention estimation, for human motion prediction with moving obstacles. While these solutions worked effectively on real-world systems, they rely significantly on the optimisation of their model parameters for each separate pair of interactions, and on the clustering methods to represent the observed scene.

In~\cite{antonucci2021efficient}, the authors proposed real-time human motion prediction for robotics applications using physics-informed neural networks that embed SFM dynamic equations. Their model was trained on synthetic data and validated offline on one person only, without any domain-shift tests. While this latter is perhaps the closest to our current work, we deploy and evaluate both neural-only and neuro-symbolic approaches for context-aware human motion prediction considering multiple moving agents and walking patterns.

%% file: approach.tex
\section{neuROSym Architecture} \label{sec:appr}

In our previous work~\cite{mghames2023neuro}, we proposed a neuro-symbolic architecture for context-aware human motion prediction and validated its accuracy performance against baseline models in an offline setting, where public datasets were locally stored and used to train and test offline the models under investigation. It is well known, however, that the prediction performance may degrade when pre-trained models are transferred to real-world settings, especially due to domain-shift changes. Therefore, in this section, we present the~\emph{neuROSym} package for deployment and validation on real robots, which extends our previous work and provides an online evaluation tool, in terms of accuracy and runtime performance, for neural-only and neuro-symbolic prediction models.

\subsection{Main neuROSym Components}
The new ROS package~\emph{neuROSym}, illustrated in Fig.~\ref{fig:intro}, consists of the following three nodes:

\begin{itemize}
    \item \textbf{Inference model node}: it implements two subscribers to the same observational data topic whose messages are generated by a human tracker library. In parallel, it implements two publishers for the data visualisation and analytics node. Each pair subscriber-publisher corresponds to either ground truth or predicted samples. The node implements also the inference model for the prediction method under investigation. In this paper, we benchmark two state-of-the-art models: SGAN~\cite{gupta2018social} and NeuroSyM~\cite{mghames2023neuro}. We re-trained both models on two public datasets: Zara01 from the UCY dataset~\cite{lerner2007crowds} and, for the first time, on the THOR\footnote{\url{http://thor.oru.se/}} dataset~\cite{thorDataset2019}, where human motion patterns differ from the previous one.
    
    \item \textbf{Data visualisation and analytics node}: this node runs in parallel to the inference node in order to generate, online, plots of the ground truth and predicted trajectories. It also generates average performance metrics, simultaneously to the visual plots.
    
    \item \textbf{Data post-processing node}: this node is required to perform corrections in case the human tracking system misses some detections. If that happens, some people would be assigned different IDs over time. This node uses an offline ROS-Rviz visualiser to help matching different IDs of the same person in the scene.
\end{itemize}

\subsection{Inference Model} 
\begin{figure}
    \centering
    \includegraphics[trim={0cm 1cm 12.5cm 0cm}, clip, scale=0.55]{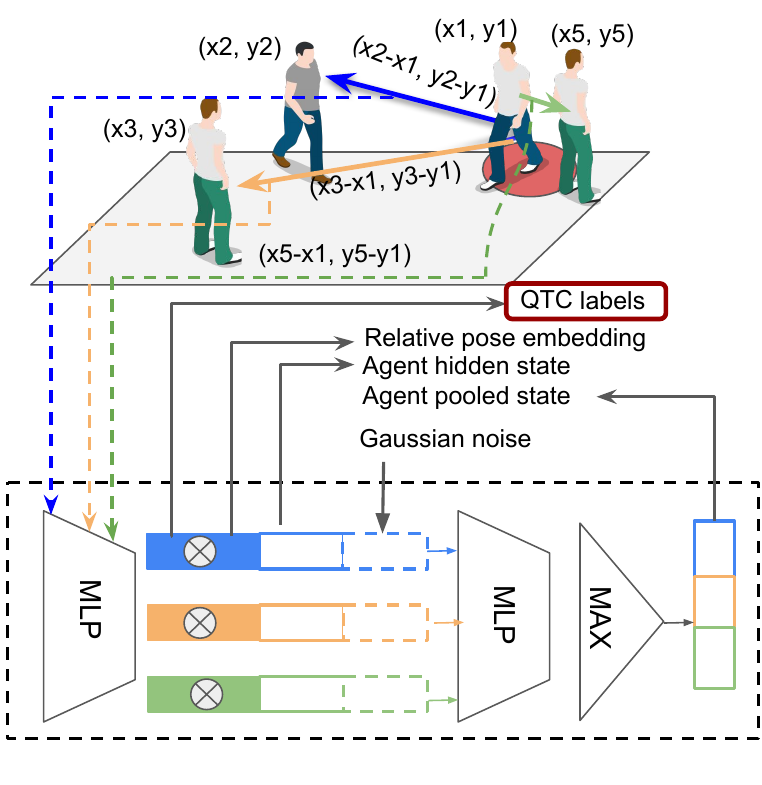}
    \caption{The NeuroSyM pooling mechanism with prior QTC knowledge injected into the output of the relative pose embedding layer.}
    \label{fig:sgan}
\end{figure}

%While the neuro-symbolic model of the SGAN architecture (NeuroSyM SGAN) leverages on a generator-discriminator component, the inference model of the NeuroSyM architecture, as well as of the neural-only architecture (SGAN) leverages only on the generator component.
The inference model node of \emph{neuROSym} is based on our previous work~\cite{mghames2023neuro}.
Here the generator part of both the NeuroSyM and the SGAN architectures consists of an encoder-pooling-decoder set of layers. In the following, we present briefly how the pooling mechanism of NeurSyM-SGAN (Fig.~\ref{fig:sgan}) incorporates the symbolic QTC knowledge in one of its layers. For a detailed explanation of the NeuroSyM architecture, including its performance on desktop experiments, we remand to the original paper~\cite{mghames2023neuro}. 
%but for the sake of this paper we report in table~\ref{fig:sgan_bars} a summary of the results in~\cite{mghames2023neuro}.

\begin{figure}\centering
\includegraphics[trim={0cm 0cm 0cm 0cm},clip, scale=0.22]{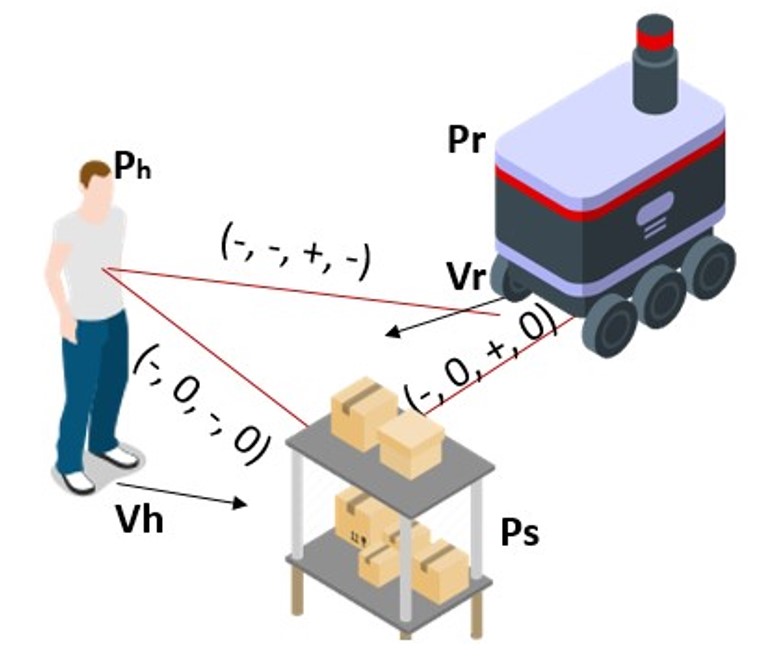}
\caption{An example of QTC$_{C1}$ representation of interactions between three body points $P_h$, $P_r$, and $P_s$.}\label{fig:qtcc}
\end{figure}

\textbf{Qualitative formulation of spatial interactions.} 
A spatial interaction is represented by a vector of $m$ QTC relations~\cite{delafontaine2012qualitative}, which consist of qualitative symbols ${q_i\in\{-, 0, +\}}$, for $i = 1, ..., m$. Among the different QTC versions, NeuroSyM adopts the double-cross QTC$_{C1}$, since it better represents the dynamics of the agents in our application scenario. More specifically, NeuroSyM was tested in~\cite{mghames2023neuro} with the four symbols $\{q_1, \, q_2, \, q_3, \, q_4\}$, where $q_1$ and $q_2$ represent the relative motion between a pair of agents (moving towards or away from), while $q_3$ and $q_4$ represent the side relation (moving to the left or to the right of). An example of QTC$_{C1}$ relations is shown in Fig.~\ref{fig:qtcc}. %The $QTC_{C1}$ type is illustrated in Fig.~\ref{fig:qtcc} for a case of interaction between three body points. 

%Given the time series of two moving points, $P_h$ and $P_r$, the qualitative interaction between them is expressed by the symbols $q_i$ as follows: 
\begin{comment}
\begin{small}
\centering
\begin{align*}
(q_1)  ~ &- : d(P_h|t^-, P_r|t) > d(P_h|t, P_r|t) \\
&0: d(P_r|t^-, P_r|t) = d(P_h|t, P_r|t) \\
 &+: d(P_h|t^-, P_r|t) < d(P_h|t, P_r|t) \\
(q_2) ~ &\text{same as $q_1$, but swapping $P_h$ and $P_r$} \\
(q_3) ~ &-: \|\vec{P_h^{t^+}P_h^t} \wedge \vec{P_r^tP_h^t}\| < 0 \\
 &0: \|\vec{P_h^{t^+}P_h^t} \wedge \vec{P_r^tP_h^t}\| = 0 \\
&+: \text{all other cases} \\
(q_4) ~ &\text{same as $q_3$, but swapping $P_h$ and $P_r$} 
\end{align*}
\end{small}

\noindent where $d(.)$ is the euclidean distance between two positions, and $\wedge$ is the cross-product notation between two vectors. 
\end{comment}

\textbf{Neuro-symbolic architecture.} 
%The neuro-symbolic approach for motion prediction (NeuroSyM) can be implemented to every related work in the field to enhance the predictive motion performance. For the time being,
To label interactions, NeuroSyM exploits QTC$_{C1}$ and the related concept of Conceptual Neighbourhood Diagram~(CND) described in~\cite{delafontaine2012qualitative,van2005conceptual}. 
The nodes of a CND are different QTC states, while edges represent the ``closeness'' of two QTC states at time $t$ and $t+1$.
%\noindent where, for practical reasons, the symbols ``+'' and ``-'' are associated to the numerical values ``+1'' and ``-1'', as in~\cite{delafontaine2012qualitative}. 
In~\cite{mghames2023neuro}, we formulated the interaction label $\alpha_{CND}$ for each QTC state as the likelihood of a transition in the CND as follows:
\begin{equation} \label{eq:labeling}
    \alpha_{CND} = P(\mathrm{QTC}_{t+1} | \mathrm{QTC}_t) = \frac{1}{N_{Tr}}
\end{equation}
where $N_{Tr}$ is the number of possible transitions from the current state.
In practice, $\alpha_{CND}$ represents the level of stability, or reliability, of a QTC state. The higher the number of possible neighbour states, the lower the likelihood to transition into one of them. Given an interaction at time $t$, we associate its label to the next one (observed or predicted) at $t+1$. Typically, an interaction between agents A and B is calculated as an embedding of their relative pose as follows:
\begin{equation}\label{eq:inter}
     I_{AB} = \emph{Dense} (X_B-X_A)
\end{equation}
where $Dense(\cdot)$ is a fully connected layer. The symbolic processing transforms Eq.~\ref{eq:inter} into $\alpha_{CND}~I_{AB}$.
The QTC knowledge is domain-agnostic and therefore it can be applied to any neural network for time-series data modeling and prediction.

\begin{comment}
\begin{figure} 
    \centering
     \includegraphics[trim={0cm 0cm 0cm 0cm},scale=0.35]{dwg/neurosym_results.png}
    \caption{A bar chart summarizing the results obtained in~\cite{mghames2023neuro} for the mean relative gain (in \%) of NeuroSyM architecture compared to its baseline one, in terms of ADE and FDE accuracy metrics over the 5 open-source UCY-ETH datasets and over the medium and long term prediction horizons.}
    \label{fig:sgan_bars}
\end{figure}
\end{comment}

%% file: experiments.tex
\section{Experiments} \label{sec:exp}

\subsection{Experimental Setup}  
\begin{figure*}
    \centering
    \begin{subfigure}[c]{0.3\textwidth}
    \centering
    \includegraphics[trim={4cm 0cm 0cm 0cm},scale=0.0245]{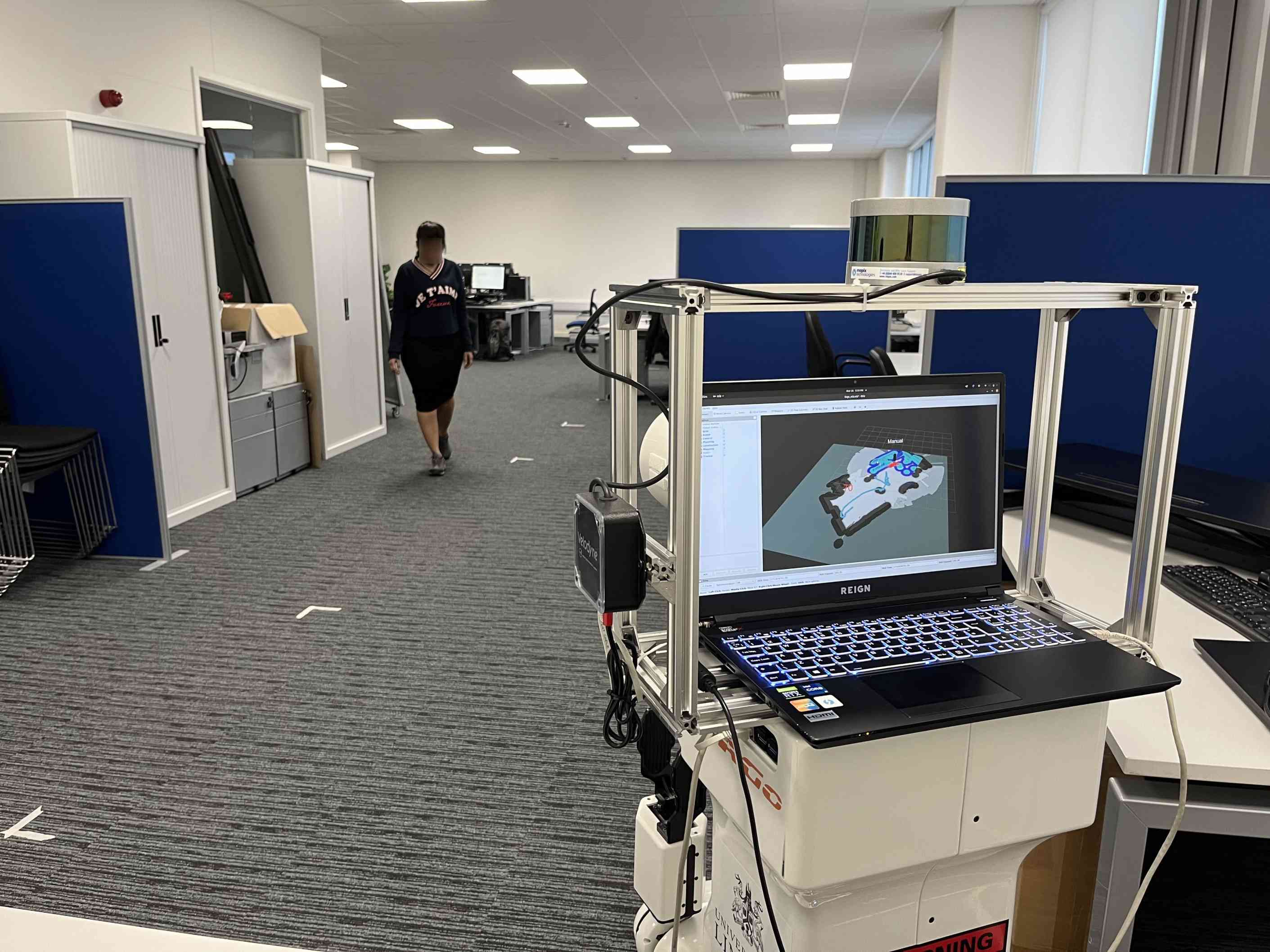}
    \caption{}
    \label{fig:exp1}
    \end{subfigure}
    \hspace{15pt}
    \begin{subfigure}[c]{0.31\textwidth}
       \centering
    \includegraphics[trim={0cm 0cm 0cm 50cm},clip, scale=0.034]{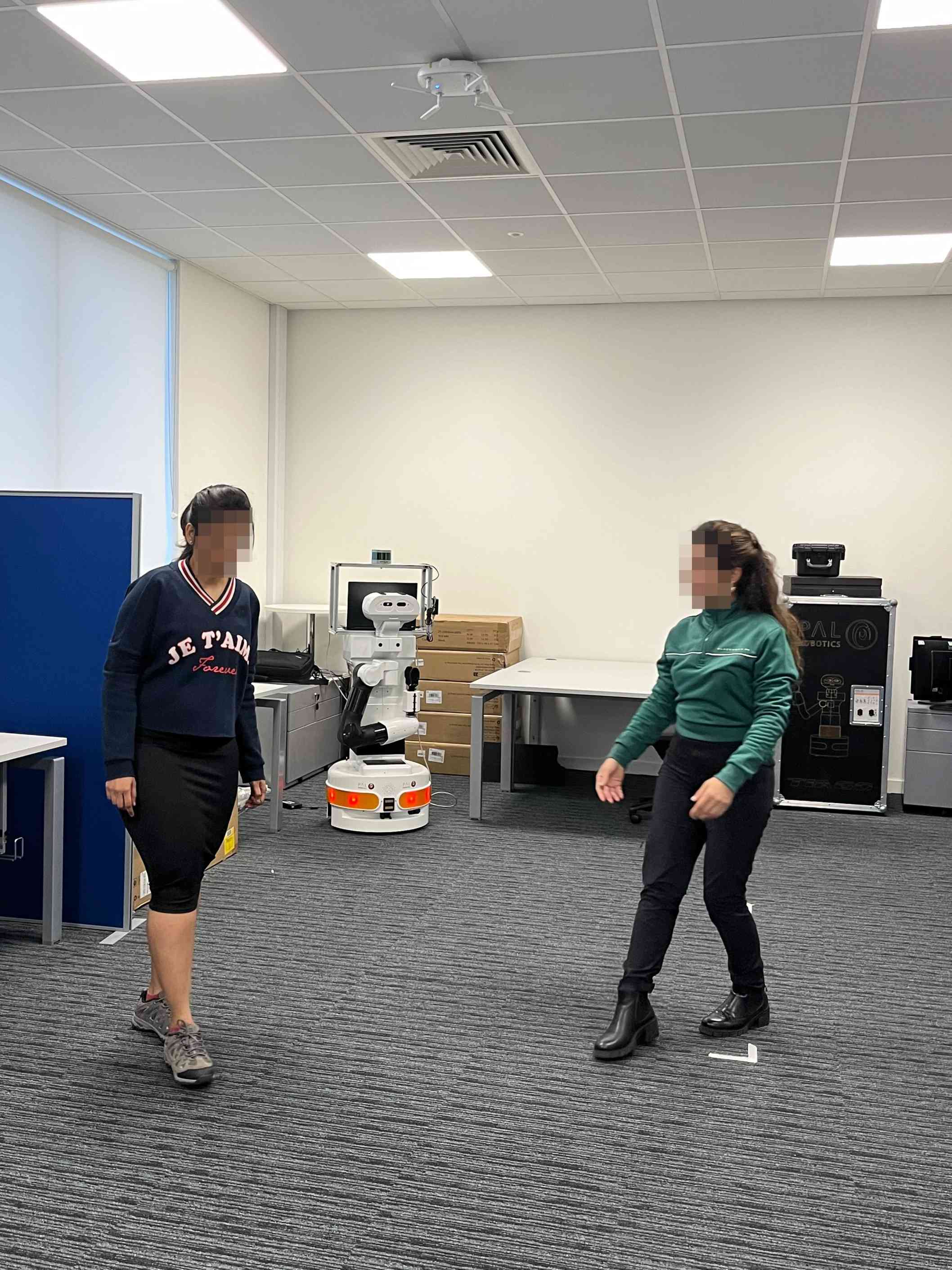}
    \caption{}
    \label{fig:exp2}
    \end{subfigure}
     %\hspace{10pt}
    \begin{subfigure}[c]{0.31\textwidth}
    \centering
    \includegraphics[ trim={0cm 0cm 10cm 0cm}, clip, scale=0.23]{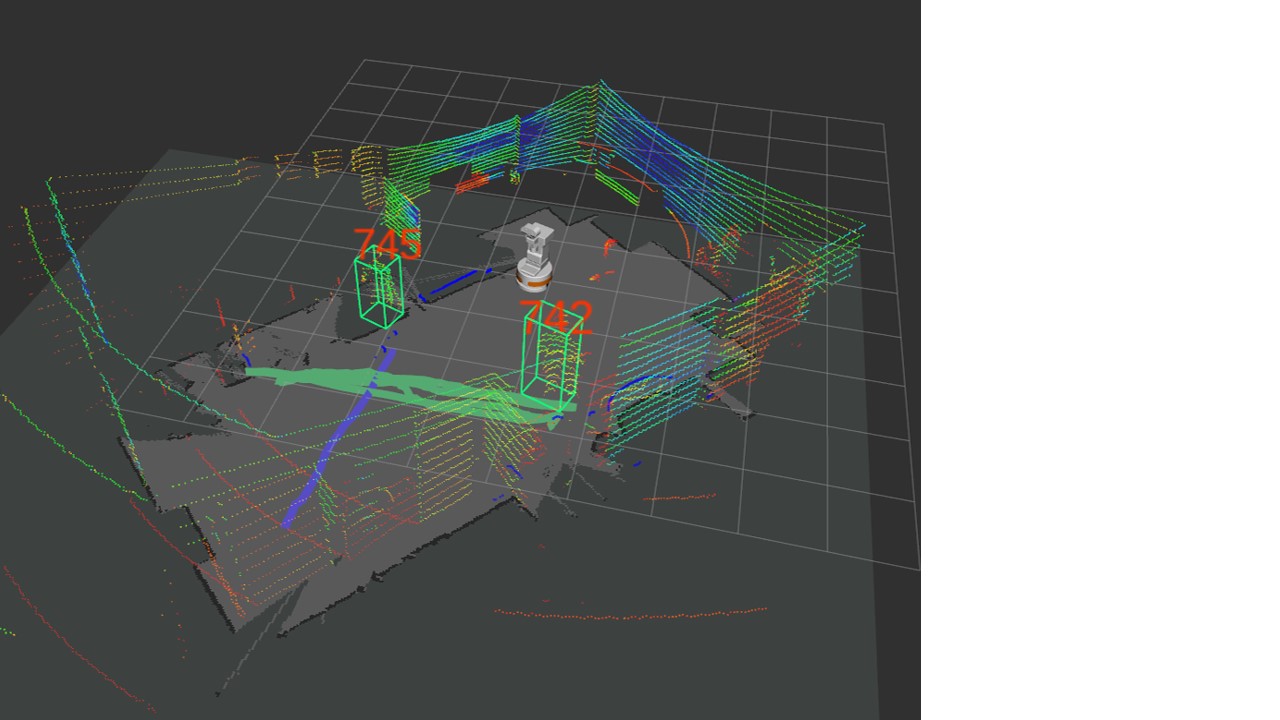}
    \caption{}
    \label{fig:rviz}
    \end{subfigure}
    \caption{(a) Experimental Scenario A with two humans walking parallel to each other towards their goal (room end) and back, repetitively. The online trajectory prediction is performed by models trained on the UCY-Zara01 dataset. (b) Experimental Scenario B with two humans crossing each other's path. Here the models are trained on the THOR dataset. (c) RViz visualization of the Bayes People Tracker with human bounding-boxes extracted from the robot's LiDAR point-clouds.}
    \label{fig:scenario}
\end{figure*}

We used a TIAGo\footnote{\url{https://pal-robotics.com/robots/tiago/}} mobile robot to monitor the motion of two people over a time period of 2 minutes. The robot was positioned at the corner of the experimental room (5m~$\times$~8.2m) and was equipped with a Velodyne VLP-16 3D LiDAR sensor, as shown in Fig.~\ref{fig:exp1}. This LiDAR features 16 scan channels, providing $360^{\circ}$ horizontal and $30^{\circ}$ vertical fields-of-view. The robot's torso was set at the minimum height of approximately $1.2$m to maximise the LiDAR's chance of detecting nearby individuals. In order to track people in the scene, we run a Bayes People Tracker\footnote{\url{https://github.com/LCAS/bayestracking}}~\cite{yan2017online} using point-cloud data from the LiDAR at a frequency of 10Hz.
Fig.~\ref{fig:rviz} shows an RViz screenshot with two humans tracked by the robot.

We conducted two types of experiments, illustrated in Figs.~\ref{fig:exp1}-\ref{fig:exp2}. During these, we recorded the runtime of each inference model (SGAN baseline and NeuroSyM). We registered the rosbag file of each experiment (four in total) for offline processing. The system was running on a computer with 11th Gen Intel® Core™ i7-11800H processor and NVIDIA GeForce RTX 3080 16GB GPU. The two experiment settings are described next.

\textbf{Scenario A: ``all-forward'' motion behaviour.}
Both SGAN and NeuroSyM were trained on the UCY-Zara01 pedestrians dataset~\cite{lerner2007crowds}. The UCY dataset consists of real pedestrian trajectories with rich multi-human interactions captured at 2.5Hz. It includes three sequences (Zara01, Zara02, and UCY), taken in public spaces from top-view videos. The motion pattern of the pedestrians resembles the \emph{all-forward} motion pattern replicated in our experiments (i.e. people walking in parallel directions) and illustrated in Fig.~\ref{fig:exp1}.

\begin{figure}[t!]
        \centering
%\begin{minipage}[c]{.55\columnwidth}
%    \begin{subfigure}[t]{\columnwidth}
%        \includegraphics[trim={14cm 1cm 12cm 1cm}, clip, scale=0.22]{dwg/luca_obs_traj_thor_bag1_16.png}
%        %\caption{}
%    \end{subfigure}
%\end{minipage}
%\begin{minipage}[c]{.4\columnwidth}
%    \begin{subfigure}[c]{\columnwidth}
%        \includegraphics[width=\columnwidth]{dwg/frame_80.png}
%       % \caption{}
%    \end{subfigure}
%    \begin{subfigure}[c]{\columnwidth}
%        \includegraphics[width=\columnwidth]{dwg/frame_100.png}
%        %\caption{}
%    \end{subfigure}
%    \begin{subfigure}[c]{\columnwidth}
%        \includegraphics[width=\columnwidth]{dwg/frame_160.png}
%        %\caption{}
%    \end{subfigure}
%\end{minipage}
\includegraphics[width=0.9\columnwidth]{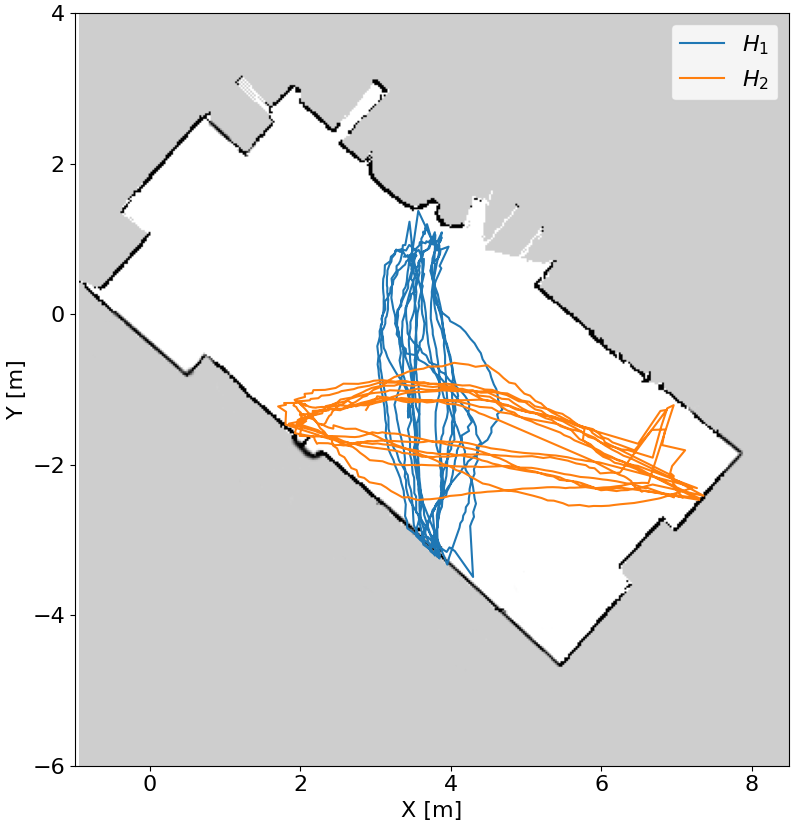}\\
\vspace{2mm}
\includegraphics[width=0.3\columnwidth]{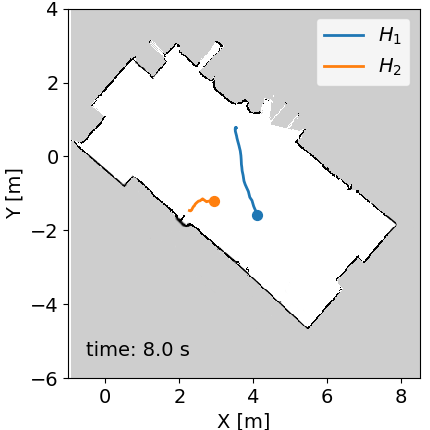}
\includegraphics[width=0.3\columnwidth]{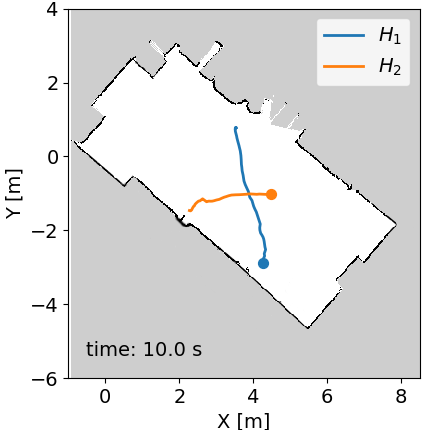}
\includegraphics[width=0.3\columnwidth]{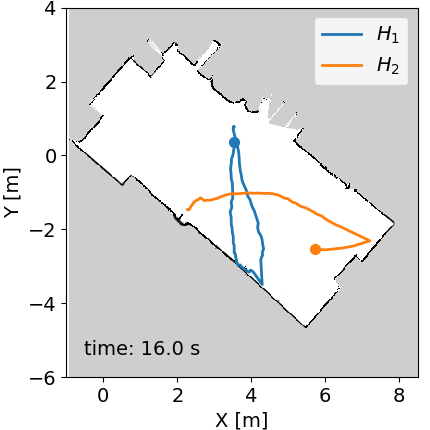}
\caption{(Top) Full human motion trajectories in Scenario B, where two people $H_1$ and $H_2$ (dynamic objects) move back and forth to their destinations (static objects), crossing each other's path and avoiding collisions. (Bottom) Snapshots at frames t = 8, 10, and 16s, from left to right, respectively.}\label{fig:trajs}
\end{figure}

\textbf{Scenario B: ``cross-path'' motion behaviour.}
The inference models were trained on the THOR dataset~\cite{thorDataset2019}, which was recorded by the authors with a motion capture system at 100Hz. This indoor dataset contains motion interactions among people and their environment, including avoidance of static and dynamic obstacles (e.g. humans, robot, static objects) by people trying to reach their goal locations. Similar motion patterns were replicated in our \emph{cross-path} scenario, as illustrated in Figs.~\ref{fig:exp2} and~\ref{fig:trajs}.

\begin{comment}
\begin{figure}[t]\centering
\includegraphics[trim={10cm 1cm 10cm 1cm}, clip, scale=0.28]{dwg/luca_obs_traj_thor_bag1_16.png}
\caption{Human motion trajectories in Scenario B, where two people $H_1$ and $H_2$ move back and forth to their destinations, crossing each other's path and avoiding collisions.} \label{fig:trajs}
\end{figure}
\end{comment}

\subsection{Data Processing}
%Following a visual inspection through Rviz, we noticed that the tracker misses some frames, e.g. when a person is partially hidden by the passage of another person. The tracker therefore assigns a new ID to the same person after he/she got uncovered. The latter has been approached with a deep association metric in~\cite{wojke2017simple} leveraging RGB data. This approach will be the direction of our future work for wide field-of-view sensors as the LiDAR and Fisheye camera. To solve the tracker issue, we conducted an offline post-processing of the rosbags and people trajectories. We identified the IDs that belong to the same person through Rviz and we republished the trajectory messages on a new ROS topic including the messages with empty frames. Two strategies can be adopted to deal with the missed frames (i.e. empty ROS messages): (a) interpolation, (b) removing the person for which at least one sample in the observed sequence has empty attributes (i.e. coordinates). Since the interpolation cannot generalise well to the case where people leave the field-of-view but can perform well when people are within the scene and occluded, we adopted the strategy of removing the pedestrian in question from the observed sequence

The ROS inference node processes the data sequentially, with an observed time window of 8 samples. %while filtering out each person's trajectory from the corresponding sequence if a sample among the 8 (for publishing the predicted trajectory messages) or among the 16 (for publishing the ground truth trajectory messages) has empty attributes.
Human trajectories affected by tracking errors (e.g. because of occlusions) were filtered out and not considered.
The ROS inference node and the visualisation node run simultaneously, showing predicted and ground-truth trajectories at runtime.  
%Driven by the variability in people's motion pattern and the tracker repeatability issue for the same experimental setting, 
The performance comparison between the baseline SGAN model and the NeuroSyM architecture was conducted on the recorded rosbag files.

\subsection{Results and Discussion}  
We evaluated the average accuracy of each inference model over the 2-minutes sessions of both experimental settings. The results are reported in Tables~\ref{tab:acc1} and~\ref{tab:acc2}, which include average displacement error~(ADE) and final displacement error~(FDE). These tables present 8 results in total, 4 for each scenario (A and B).
%The latter is because we replayed offline each rosbag for each single inference model under consideration.
%In addition to ADE and FDE, Tables~\ref{tab:acc1} and~\ref{tab:acc2} report also the total number of trajectories predicted in the 2-minutes interval, the total number of trajectories predicted with at least two people detected.
%the ADE and FDE relative gains.
We can see that, in all of the four cases, the higher accuracy achieved by NeuroSyM significantly reduced both ADE and FDE values compared to the SGAN baseline.

\begin{table}
\begin{center}
\begin{tabular}{c|*2c|*2c}
\bf{Scenario A} & \multicolumn{2}{c|}{\bf{rosbag 1}} & \multicolumn{2}{c}{\bf{rosbag 2}}\\
\hline
 & \textbf{SGAN} &  \textbf{NeuroSyM} & \textbf{SGAN} &  \textbf{NeuroSyM} \\
\hline
\textbf{Avg. ADE }(m) &  12.4 &  \bf{7.06} &  16.32 &  \bf{2.52} \\
\textbf{Avg. FDE }(m) & 2.28 & \bf{1.31} & 3.24 & \bf{0.68} \\
%\textbf{Total Seq.} &  6 & 6 & 7* & 3 \vspace{-3pt}\\ 
%\textbf{(with ppl $\geq$ 2)} &  (4) & (3) & (5) & (3) \\
%\hline
%\textbf{Seq. with ppl $\geq$ 2} & I, bag-1 & 4 & 3\\
%\hline
% \textbf{ADE Rel. Gain }(\%) & I, bag-1 & - &  +43.06\\
%\hline
% \textbf{FDE Rel. Gain }(\%)  & I, bag-1 & - & +42.54 \\
%\hline
%\hline
%\bf{Scenario A, rosbag 2} & \textbf{Baseline} &  \textbf{NeuroSyM}  \\
%\hline
%\textbf{Avg. ADE }(m) &  16.32 &  2.52 \\
%\hline
%\textbf{Avg. FDE }(m) & 3.24 & 0.68 \\
%\hline
%\textbf{Total Seq.} & $7^*$ & 3  \\
%\hline
%\textbf{Seq. with ppl $\geq$ 2} & 5 & 3\\
%\hline
% \textbf{ADE Rel. Gain }(\%) & I, bag-2 & - & +84.55 \\
%\hline
% \textbf{FDE Rel. Gain }(\%)  & I, bag-2 & - & +79.01 \\
%\hline
\end{tabular}
\caption{Accuracy evaluation for Scenario A, in terms of average displacement error~(ADE) and final displacement error~(FDE), over 2-minutes long experiments. 
%The note (*) highlights that, for the experimental setting in question, over 3 sequences only the average ADE = 17.87 (m) and the average FDE = 2.98 (m).
} \label{tab:acc1}
\end{center}
\end{table}

\begin{table}
\begin{center}
\begin{tabular}{c|*2c|*2c}
\bf{Scenario B} & \multicolumn{2}{c|}{\bf{rosbag 1}} & \multicolumn{2}{c}{\bf{rosbag 2}}\\
\hline
 & \textbf{SGAN} &  \textbf{NeuroSyM} & \textbf{SGAN} &  \textbf{NeuroSyM} \\
\hline
\textbf{Avg. ADE }(m) &  10.88 &  \bf{5.7} &  24.27 &  \bf{9.87} \\
\textbf{Avg. FDE }(m) & 2.67 & \bf{1.4} & 5 & \bf{1.8} \\
%\textbf{Total Seq.} &  3 & 3 & 5 & 5 \vspace{-3pt}\\ 
%\textbf{(with ppl $\geq$ 2)} &  (1) & (1) & (3) & (3) \\
%\hline
%Features & \textbf{Scenario} & \textbf{Baseline} &  \textbf{NeuroSyM}  \\
%\hline
%\textbf{Avg. ADE }(m) & II, bag-1 &  10.88 &  5.7 \\
%\hline
%\textbf{Avg. FDE }(m) & II, bag-1 & 2.67 & 1.4 \\
%\hline
%\textbf{Total Seq.} & II, bag-1 &  3 & 3  \\
%\hline
%\textbf{Seq. with ppl $\geq$ 2} & II, bag-1 & 1 & 1\\
%\hline
% \textbf{ADE Rel. Gain }(\%) & II, bag-1 & - & +47.6 \\
%\hline
% \textbf{FDE Rel. Gain }(\%)  & II, bag-1 & - & +47.56 \\
%\hline
%\hline
%\textbf{Avg. ADE }(m) & II, bag-2 &  24.27 &  9.87 \\
%\hline
%\textbf{Avg. FDE }(m) & II, bag-2 & 5 & 1.8 \\
%\hline
%\textbf{Total Seq.} & II, bag-2 &  5 & 5  \\
%\hline
%\textbf{Seq. with ppl $\geq$ 2} & II, bag-2 & 3 & 3\\
%\hline
% \textbf{ADE Rel. Gain }(\%) & II, bag-2 & - & +59.3 \\
%\hline
% \textbf{FDE Rel. Gain }(\%)  & II, bag-2 & - & +64 \\
%\hline
\end{tabular}
\caption{Accuracy evaluation for Scenario B, in terms of average displacement error~(ADE) and final displacement error~(FDE), over 2-minutes long experiments.} \label{tab:acc2}
\end{center}
\end{table}

\begin{figure}[t!]
    \centering
%    \begin{subfigure}[c]{\columnwidth}
%    \centering
    \includegraphics[width=0.95\columnwidth]{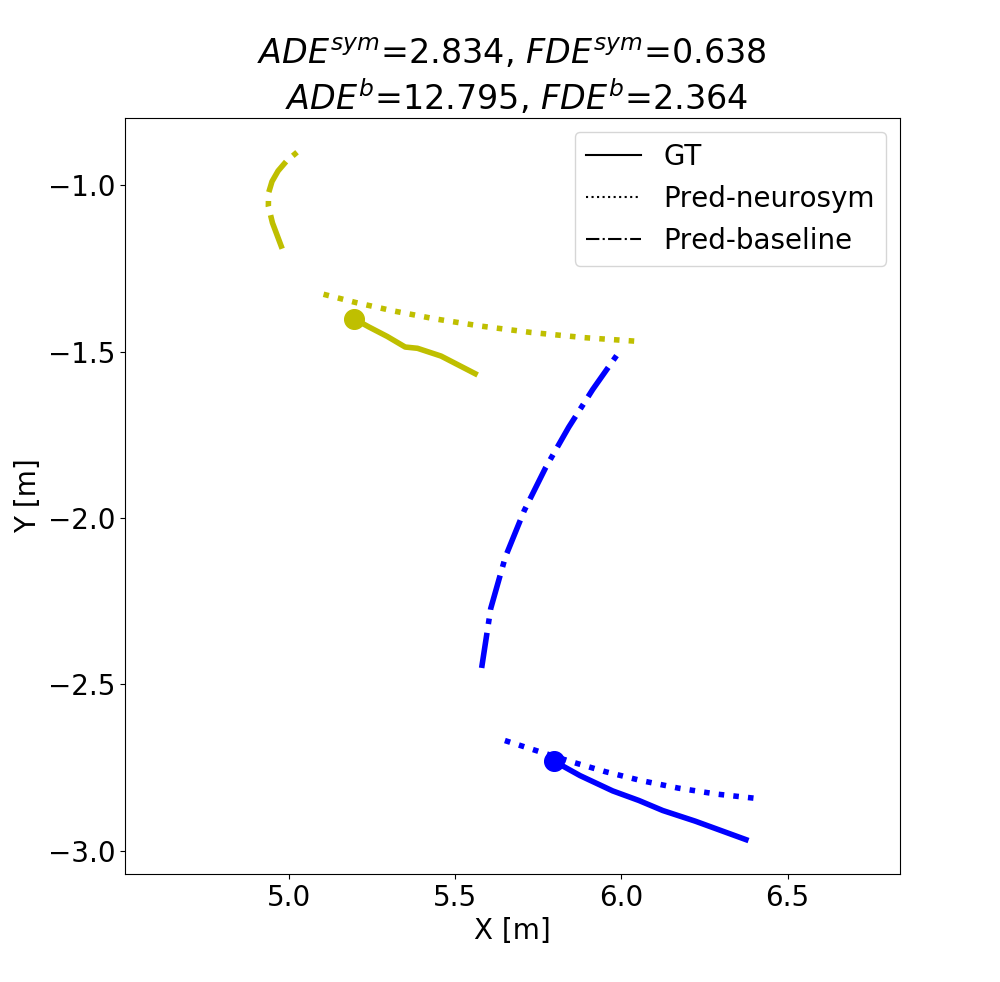}\\\vfil
%    \caption*{}
%    \label{fig:bag1_baseline}
%    \end{subfigure}
%    \begin{subfigure}[c]{\columnwidth}
%      \centering
    \includegraphics[width=0.95\columnwidth]{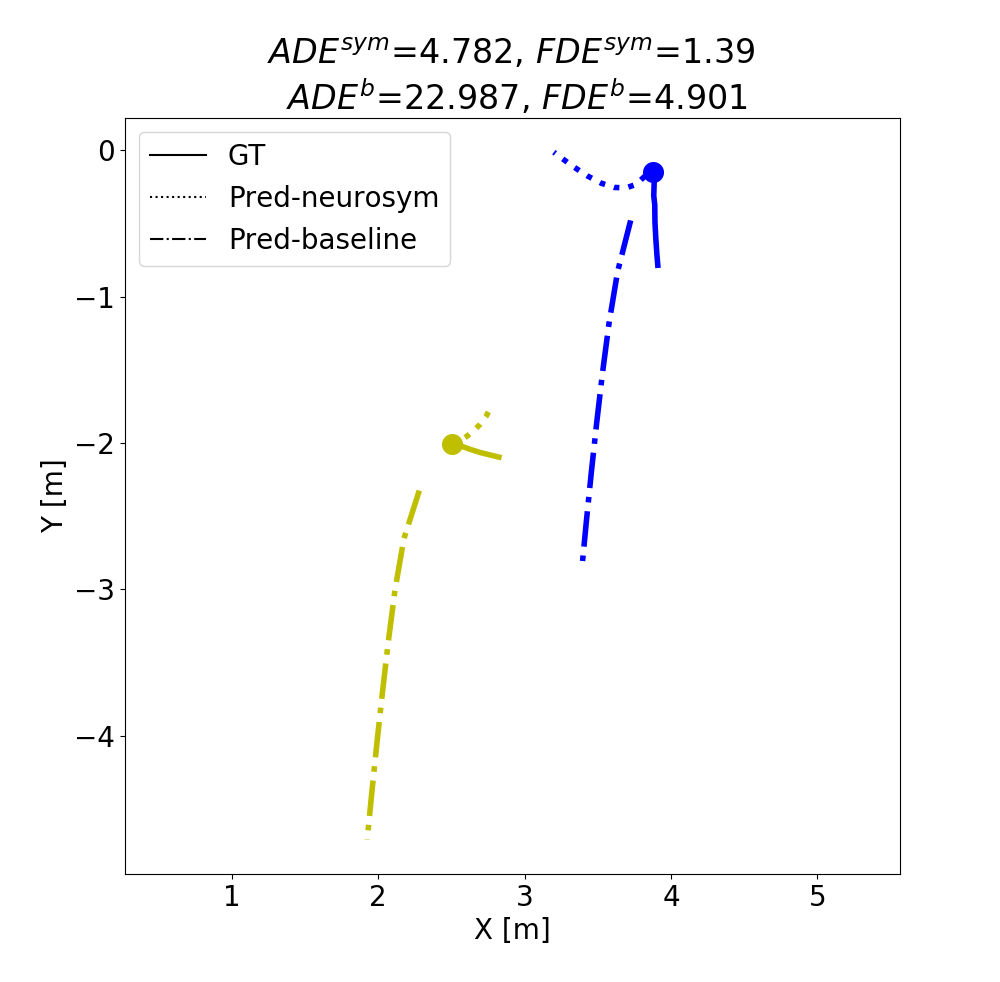}
%    \caption*{}
%    \label{fig:bag1_neurosym}
%    \end{subfigure}
%    \vspace{-18pt}
    \caption{Examples of trajectories of two people for a single sequence, extracted from the total number of sequences generated within a 2-minutes experiment. The solid lines are the ground-truth trajectories of the predictions (8 samples each), while the dotted and the dash-dotted lines are the neuro-symbolic and the baseline predictions, respectively. The small circle is the starting point. (Top) Experimental scenario A. (Bottom) Experimental scenario B. Superscripts \emph{sym} and \emph{b} denote neuro-symbolic and baseline, respectively.}
    \label{fig:predictions}
\end{figure}

Fig.~\ref{fig:predictions} illustrates some examples of ground truth and predicted trajectories in Scenarios~A (top plot) and B (bottom plot), with the corresponding ADE and FDE metrics. We can clearly see that ADE and FDE are lower, in both plots, where the NeuroSyM model was used.

We also evaluated the average runtime of each inference model in both experimental scenarios. The results are reported in Table~\ref{tab:time}, showing that NeuroSyM model is slightly slower than, but still comparable to, the SGAN baseline.
%We note that, in the more complex scenario (scenario B), the maximum inference runtime reached $\sim$ 7s for the NeuroSyM model deployment compared to $\sim$ 5s for its baseline architecture. Also, the runtime results are reported starting from the collection of the observation samples.

\begin{comment}
\begin{table}
\begin{center}
\begin{tabular}{c|c|c|c}
\hline
\textbf{Scenario} & \textbf{Model} & \textbf{Average time} (s) & \textbf{Rel. Gain} (\%) \\
\hline
I & SGAN & 4.17 & +22.3\\
\hline
I & NeuroSyM SGAN & 5.37  & -\\
\hline
II & SGAN & 5.19 & +29.5 \\
\hline
II & NeuroSyM SGAN & 7.36  & -\\
\end{tabular}
\caption{Runtime evaluation over 2-minutes experimental duration and data recordings. } \label{tab:time}
\end{center}
\end{table}
\end{comment}

\begin{table}
\begin{center}
\begin{tabular}{c|c|cc}
 & \textbf{Scenario} & \textbf{SGAN} & \textbf{NeuroSyM} \\
\hline
\multirow{2}{*}{\textbf{Average time} (s)}
  & \bf{A} & \bf{4.17} &  5.37\\
%\hline
%\textbf{Rel. Gain} (\%) & A & +22.3 & -  \\
%\hline
%\hline
  & \bf{B} & \bf{5.19} & 7.36 \\
%\hline
%\textbf{Rel. Gain} (\%) & II & +29.5  & -\\
\end{tabular}
\caption{Runtime evaluation for the two scenarios.} \label{tab:time}
\end{center}
\end{table}

From Tables~\ref{tab:acc1},~\ref{tab:acc2}, and~\ref{tab:time}, we can conclude that, although the NeuroSyM architecture requires more time to predict human trajectories compared to the SGAN baseline, it is still relatively fast and, with some code optimisation, suitable for real-time deployment. In particular, the trade-off between runtime and accuracy is clearly in favour of the NeuroSyM solution, since its QTC-based context-awareness enables more accurate motion predictions.

%% file: conclusion.tex
\section{Conclusion} \label{sec:conc}
In this work, we implemented and deployed a ROS-based architecture, called \emph{neuROSym}, for neural-only and neuro-symbolic motion prediction on real-world robotic systems. Using this framework, we experimentally evaluated the accuracy and runtime performance of two predictions models, SGAN and NeuroSyM, during online inference in two scenarios with different human motion patterns. The results show a trade-off between accuracy and runtime performance in favor of the NeuroSyM solution, which is particularly suitable for human-aware robot navigation. 
%and in potential plant facilities management. 
Our future work will extend the evaluation of \emph{neuROSym} to more diverse and challenging scenarios, including complex human motion patterns with multiple people. We will also consider more robust people trackers and test against different baseline architectures with static- and dynamic-context awareness.
%, possibly extending previous work in this area~\cite{wojke2017simple} to associate detections based on semantics. Finally, we will further evaluate our methods on more diverse and challenging scenarios with complex motion patterns, and against different baseline architectures with static-context awareness. 